\documentclass[letterpaper]{article} % DO NOT CHANGE THIS
\usepackage{aaai2026}  % DO NOT CHANGE THIS
\usepackage{times}  % DO NOT CHANGE THIS
\usepackage{helvet}  % DO NOT CHANGE THIS
\usepackage{courier}  % DO NOT CHANGE THIS
\usepackage[hyphens]{url}  % DO NOT CHANGE THIS
\usepackage{graphicx} % DO NOT CHANGE THIS
\usepackage{natbib}  % DO NOT CHANGE THIS AND DO NOT ADD ANY OPTIONS TO IT
\usepackage{caption} % DO NOT CHANGE THIS AND DO NOT ADD ANY OPTIONS TO IT
\usepackage{algorithm}
\usepackage{algorithmic}
\usepackage{amsmath}
\usepackage{amsfonts}

\usepackage{newfloat}
\usepackage{listings}
\DeclareCaptionStyle{ruled}{labelfont=normalfont,labelsep=colon,strut=off} % DO NOT CHANGE THIS
\floatstyle{ruled}
\newfloat{listing}{tb}{lst}{}
\floatname{listing}{Listing}
\title{Radiation-Preserving Selective Imaging for Pediatric Hip Dysplasia:\\
A Cross-Modal Ultrasound-Xray Policy with Limited Labels}
\author{
    Duncan Stothers\textsuperscript{\rm 1,3}\thanks{Corresponding author: duncanstothers@alumni.harvard.edu},
    Ben Stothers\textsuperscript{\rm 1,2},
    Emily Schaeffer\textsuperscript{\rm 1,2},
    Kishore Mulpuri\textsuperscript{\rm 1,2}
}
\affiliations{
    \textsuperscript{\rm 1}HIPPY lab\\
    \textsuperscript{\rm 2}University of British Columbia\\
    \textsuperscript{\rm 3}Harvard University\\
}

\usepackage{bibentry}
\begin{document}

\maketitle

\begin{abstract}
We study an ultrasound-first, radiation-preserving policy for developmental dysplasia of the hip (DDH) that requests an X-ray (XR) only when needed. 

We (i) pretrain modality-specific encoders (ResNet-18) with SimSiam on a large unlabelled registry (37{,}186 ultrasound; 19{,}546 radiographs), (ii) freeze the backbones and fit small, measurement-faithful heads on DDH-relevant landmarks and measurements, (iii) calibrate a one-sided conformal deferral rule on ultrasound predictions that provides finite-sample \emph{marginal} coverage guarantees under exchangeability, using a held-out calibration set. Ultrasound heads predict Graf $\alpha/\beta$ and femoral head coverage; X-ray heads predict acetabular index (AI), center-edge (CE) angle and IHDI grade. On our held-out labeled evaluation set, ultrasound measurement error is modest (e.g., $\alpha$ MAE $\approx$ 9.7$^\circ$, coverage MAE $\approx$ 14.0 percentage points), while radiographic probes achieve AI and CE MAEs of $\approx$ 7.6$^\circ$ and $\approx$ 8.9$^\circ$, respectively. The calibrated US-only policy is explored across rule families (alpha-only; alpha OR coverage; alpha AND coverage), conformal miscoverage levels $(\delta_\alpha,\delta_{\mathrm{cov}})$, and per-utility trade-offs using decision-curve analysis. Conservative settings yield high coverage (e.g., $\sim$0.90 for $\alpha$) with near-zero US-only rates; permissive settings (e.g., alpha OR coverage at larger deltas) achieve non-zero US-only throughput with expected coverage trade-offs. 

The result is a simple, reproducible pipeline that turns limited labels into interpretable measurements and tunable selective imaging curves suitable for clinical handoff and future external validation.
\end{abstract}

\section{Background}

\paragraph{Clinical context and age-aware measurements.}
Developmental dysplasia of the hip (DDH) spans acetabular undercoverage through dislocation across infancy and early childhood. In the radiographic domain, clinically accepted measurements include the acetabular index (AI), IHDI grading, and, when ossification permits, the lateral center-edge (Wiberg) angle and Sharp’s acetabular angle \cite{tonnis1987_book_dysplasia,narayanan2015_jpo_ihdi,doski2022_clinorthop_iHDI_upgrade,sharp1961_jbjsb_acetabular_angle,wiberg1939_acta_ce_angle}. In early infancy, standardized sonography under the Graf method (reporting $\alpha/\beta$ angles and femoral head coverage) is the dominant screening and early management tool \cite{graf1980_archorthop_ultrasound_orig,graf2006_book_hip_sonography,zieger1986_pediatradiol_us_validity,omeroglu2014_jchildorthop_us_ddh}. As ossification progresses, anteroposterior (AP) pelvic radiographs become more informative for acetabular development and surgical planning; minimizing ionizing exposure in pediatrics motivates selective XR acquisition \cite{keller2009_pediatradiol_us_vs_xray,dezateux2007_lancet_ddh_review}.

\paragraph{Ultrasound-only automation (what exists).}
A substantial body of work automates hip ultrasound tasks: standard-plane detection, quality gating, Graf typing, and direct prediction of $\alpha/\beta$ and femoral head coverage. Representative approaches include multitask CNNs that jointly detect plane adequacy and infer Graf measurements \cite{chen2022_diagnostics_graf_plane,hu2022_jbhi_ultrasound_multitask}, lightweight real-time systems designed for point-of-care guidance \cite{hsu2025_ijms_real_time_ddh_us}, and pipelines that integrate structure/landmark cues to improve robustness \cite{kinugasa2023_jpo_deep_ultrasound_ddh,chen2024_biomedj_graf_type}. Reviews of pediatric MSK ultrasound emphasize ultrasound’s safety, accessibility, and the potential for AI-assisted standardization to reduce operator dependence \cite{vanKouswijk2025_children_pedsMSK_ai}. These methods typically optimize unimodal accuracy/MAE and, when evaluated, report agreement with expert sonographers on Graf measures and coverage.

\paragraph{Radiograph-only automation (what exists).}
On AP pelvic radiographs, deep systems tackle triage, landmark detection, and explicit measurement prediction for AI, CE, and related angles. Recent efforts use local-global architectures and landmark-aware heads to increase interpretability and reduce measurement bias \cite{liu2020_tmi_misshapen_pelvis,moon2024_heliyon_auto_pelvic_params,li2019_medicine_sharp_angle_auto}. Additional work frames DDH classification and grading tasks with CNNs and detectors \cite{park2021_kjr_ddh_cnn,zhang2020_bjj_ai_ddh,fraiwan2022_bmcmi_ddh_transfer,den2023_sciprep_yolov5_ddh,xu2022_frontped_dl_system_ddh}. Systematic reviews synthesize steady progress but highlight the need for stronger labeling protocols, external validation, and reporting of clinically meaningful measurement error rather than only image-level accuracy \cite{wu2023_frontped_ai_ai_index,chen2024_josr_systematic_ai_ddh}.

\paragraph{Cross-modal US-XR learning (what remains sparse).}
Despite mature unimodal pipelines, explicitly \emph{paired} US-XR learning for DDH is scarce. Reviews note limited availability of temporally matched US-XR cohorts, inconsistent annotation schemes across modalities, and few methods that directly model the decision of whether to acquire an XR given a US \cite{wu2023_frontped_ai_ai_index,chen2024_josr_systematic_ai_ddh,vanKouswijk2025_children_pedsMSK_ai}. Emerging resources begin to address pairing and benchmarking \cite{qi2025_sciData_mtddh_dataset}, and there are early bimodal/local-global ideas in related orthopedic settings \cite{shimizu2024_sciprep_bimodal_unstable_hips}, but a practical, measurement-aware policy that trades radiation against risk with explicit guarantees remains underexplored.

\paragraph{Why the intersection matters: clinical and statistical rationale.}
US and XR encode overlapping-but not identical-morphology. Graf $\alpha$ and femoral head coverage on US reflect acetabular roof orientation and femoral containment in early infancy, while AI/CE on XR quantify acetabular slope and lateral coverage once ossification permits reliable landmarks \cite{graf1980_archorthop_ultrasound_orig,graf2006_book_hip_sonography,tonnis1987_book_dysplasia,wiberg1939_acta_ce_angle}. Clinically, a clearly normal US (e.g., high $\alpha$, adequate coverage) often reduces the immediate value of XR in very young infants, whereas abnormal or borderline US increases the value of XR to confirm severity, assess symmetry, and plan management. Statistically, this suggests a selective acquisition problem: use US-derived measurements and calibrated uncertainty to predict whether XR would meaningfully shift the diagnostic or management decision boundary.

\paragraph{From ultrasound to radiographic surrogates (US$\rightarrow$XR prediction).}
Several radiographic targets admit physiologic surrogates on ultrasound. Elevated $\alpha$ and adequate coverage reduce the likelihood of elevated AI or pathologic IHDI in age-appropriate cohorts; conversely, low $\alpha$ or poor coverage raise suspicion for acetabular underdevelopment that XR can quantify with AI/CE more precisely as ossification progresses. A measurement-aware proxy model can therefore (i) estimate XR-relevant risk from US measurements, (ii) calibrate one-sided lower bounds that certify “confidently normal” cases against clinical thresholds (e.g., $\alpha \ge 60^\circ$; coverage $\ge 50\%$), and (iii) defer to XR when uncertainty or predicted abnormality exceeds a tunable margin. This framing preserves interpretability (decisions ride on named measurements) and keeps the bridge between modalities clinically legible.

\paragraph{Decision-theoretic selective imaging and risk control.}
Selective acquisition can be cast as a utility-optimization under uncertainty: balance a (small) radiation cost against the (potentially large) cost of missing a lesion that XR would reveal. In deployment, finite-sample guarantees matter. Conformal prediction provides distribution-free \emph{marginal} coverage control for one-sided lower bounds on US measurements. We then use these bounds inside US-only rules while routing ambiguous cases to XR. Decision-curve analysis then summarizes net benefit across utility weights, exposing operating points along the safety-radiation frontier.

\paragraph{Positioning of this work.}
We contribute a pragmatic, label-efficient cross-modal policy: (1) self-supervised pretraining on large unlabeled US/XR registries with frozen backbones; (2) small, measurement-faithful heads trained on a curated set of paired studies; (3) a one-sided conformal layer that certifies “US-only” decisions against clinically standard thresholds; and (4) decision-curve analysis that makes explicit the trade between XR utilization and risk. The intersection we target, using calibrated US measurements to decide when XR adds value, addresses the practical gap highlighted by recent reviews \cite{wu2023_frontped_ai_ai_index,chen2024_josr_systematic_ai_ddh,vanKouswijk2025_children_pedsMSK_ai} and complements ongoing efforts in unimodal automation \cite{park2021_kjr_ddh_cnn,zhang2020_bjj_ai_ddh,liu2020_tmi_misshapen_pelvis,moon2024_heliyon_auto_pelvic_params,chen2022_diagnostics_graf_plane,hu2022_jbhi_ultrasound_multitask}.

\begin{figure}[!t]
  \centering
  \includegraphics[width=0.48\linewidth]{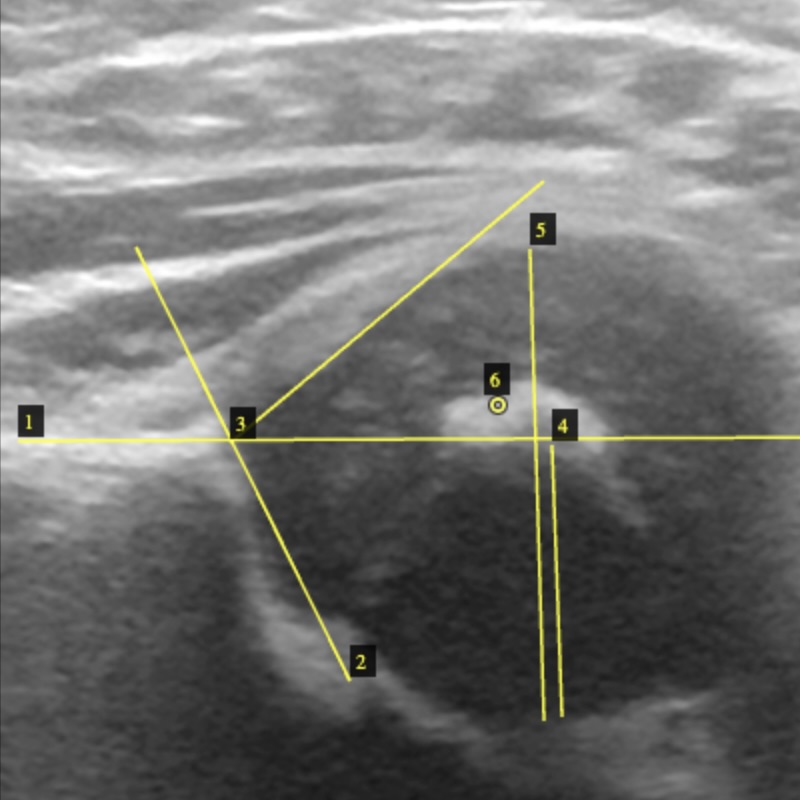}\hfill
  \includegraphics[width=0.48\linewidth]{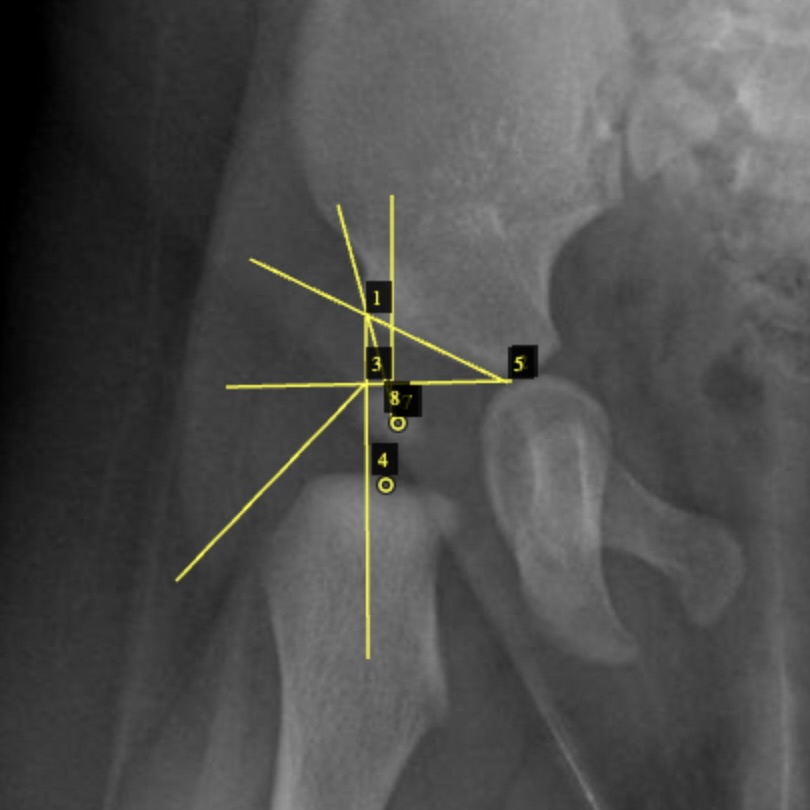}
  \caption{(Left) Pediatric hip ultrasound (US) and (Right) pediatric hip X-ray (XR). On US: annotations for the horizontal (1), Graf $\alpha$ (2), Graf $\beta$ (3), percent femoral head coverage (4/5), and location of ossific nucleus (6). On XR: P-line (1), H-line (2), 45$^\circ$ line (3), IHDI quadrant and grade (4), Acetabular Index (AI) (5), Central Edge Vertical line (6), Central Edge Angle line (CE) (7), location of ossific nucleus (8).}
  \label{fig:hips}
\end{figure}

\section{Methods}

\subsection{Data, pairing, and splits}
We assembled a large unlabeled corpus for self-supervised pretraining (US: $37{,}186$; XR: $19{,}546$ grayscale images, resized to $512{\times}512$ and channel-repeated to 3). For supervised training and policy evaluation, we curated a small, paired subset with trainee line/point annotations and strict side/date matching. We removed frog-lateral and non-AP studies with a fixed set of view and QC rules, which eliminated eight subjects. The resulting labeled set contained $321$ images from $75$ subjects. To avoid leakage, we used subject-level splits: \emph{post-train} (30 subjects; 136 images), \emph{calibration} (7 subjects; 28 images; including 26 ultrasound images), and \emph{evaluation} (38 subjects; 157 images). On \emph{evaluation}, strict matching by (subject, date, side) yielded $N{=}77$ hip pairs with at least one XR ground truth label (AI and/or CE and/or IHDI).

For each US image, the trainee recorded a horizontal baseline and $\alpha/\beta$ lines, exposed/total femoral-head lengths, and an optional ossific-nucleus point. For each XR image, the trainee recorded Hilgenreiner’s H-line, Perkin’s P-line, a $45^\circ$ reference, the H-point, an acetabular index line, and CE-angle rays, with laterality handled explicitly. Numeric targets were derived by the standard acute-angle and ratio formulas (Graf $\alpha/\beta$; coverage = exposed/total; AI = angle(H-line, AI-line); CE = angle(vertical, CE-ray); IHDI from the quadrant of the H-point).

\subsection{Self-supervised pretraining (SimSiam) and frozen encoders}
We train separate encoders for US and XR with SimSiam on the unlabeled corpora and then freeze them. Let $f_\phi$ be a ResNet-18 encoder, $h_\phi$ a projection MLP, and $q_\phi$ a prediction MLP. For an image $x$, we draw two augmentations $t_1,t_2$ to obtain views $v_1{=}t_1(x)$ and $v_2{=}t_2(x)$. Define
\[
\begin{aligned}
z_1 &= h_\phi\!\big(f_\phi(v_1)\big), \quad
z_2 = h_\phi\!\big(f_\phi(v_2)\big),\\
p_1 &= q_\phi(z_1), \quad
p_2 = q_\phi(z_2).
\end{aligned}
\]
The SimSiam loss is the stop-gradient negative cosine similarity:
\[
\mathcal{L}_{\mathrm{SSL}}(x;\phi)
= -\tfrac{1}{2}\left[ 
\frac{\langle p_1,\ \mathrm{sg}(z_2)\rangle}{\|p_1\|_2\,\|\mathrm{sg}(z_2)\|_2}
+\frac{\langle p_2,\ \mathrm{sg}(z_1)\rangle}{\|p_2\|_2\,\|\mathrm{sg}(z_1)\|_2}
\right].
\]
We train for 10 epochs per modality with standard SimSiam augmentations (random crop/flip, color jitter for XR toned down for US), then discard $h_\phi,q_\phi$ and freeze the encoder $f_\phi$ for downstream use.

\subsection{Measurement heads on frozen encoders}
Given a frozen modality-specific encoder $f_\phi$, we extract a $512$-dimensional feature $u=f_\phi(x)$ (global average pooled). We then fit small, measurement-faithful heads with task-appropriate losses.

\paragraph{Ultrasound head.}
We use a single MLP with one hidden layer (128 units) and three outputs predicting
$\hat{\alpha},\hat{\beta}\in\mathbb{R}$ (degrees) and $\widehat{\mathrm{cov}}\in\mathbb{R}$ (percentage points). Let $y^\alpha,y^\beta,y^{\mathrm{cov}}$ denote the trainee labels for a given image. The US loss is mean absolute error (MAE) with per-task scaling to balance magnitudes:
\[
\mathcal{L}_{\mathrm{US}}(x) 
= \lambda_\alpha\,\big|\hat{\alpha}-y^\alpha\big|
+ \lambda_\beta\,\big|\hat{\beta}-y^\beta\big|
+ \lambda_{\mathrm{cov}}\,\big|\widehat{\mathrm{cov}}-y^{\mathrm{cov}}\big|.
\]
In our implementation we set $\lambda_\alpha{=}\lambda_\beta{=}1$ and $\lambda_{\mathrm{cov}}{=}1$ (coverage reported in percentage points).

\paragraph{Radiograph heads.}
We fit angle regressors for AI and CE and, when present, an IHDI classifier. Let $\hat{a},\hat{e}\in\mathbb{R}$ be AI and CE predictions and $\hat{\pi}\in\Delta^{K-1}$ be softmax probabilities over $K$ IHDI grades. With angle labels $y^{\mathrm{AI}},y^{\mathrm{CE}}$ and (optional) IHDI one-hot $y^{\mathrm{IHDI}}$,
\[
\begin{aligned}
\mathcal{L}_{\mathrm{XR}}(x)
= \mu_{\mathrm{AI}}\,\big|\hat{a}-y^{\mathrm{AI}}\big|
+ \mu_{\mathrm{CE}}\,\big|\hat{e}-y^{\mathrm{CE}}\big| \\
+ \mu_{\mathrm{IHDI}}\,\mathrm{cross\_entropy}\!\left(\hat{\pi},y^{\mathrm{IHDI}}\right),
\end{aligned}
\]
where the crossentropy term is dropped if $y^{\mathrm{IHDI}}$ is absent. In all experiments we set $\mu_{\mathrm{AI}} = \mu_{\mathrm{CE}} = \mu_{\mathrm{IHDI}} = 1$, i.e., we use an unweighted sum of the available terms. When $y^{\mathrm{IHDI}}$ is missing, the classification term is omitted rather than reweighted. Given the limited dataset size and the relatively similar numeric scale of AI and CE errors, we did not attempt further loss reweighting. We train heads on the post-train split only using a fixed training schedule and freeze them for evaluation and policy analysis. The calibration split is reserved for affine bias correction and conformal calibration.

\begin{figure*}[t]
\centering
\includegraphics[width=0.32\linewidth]{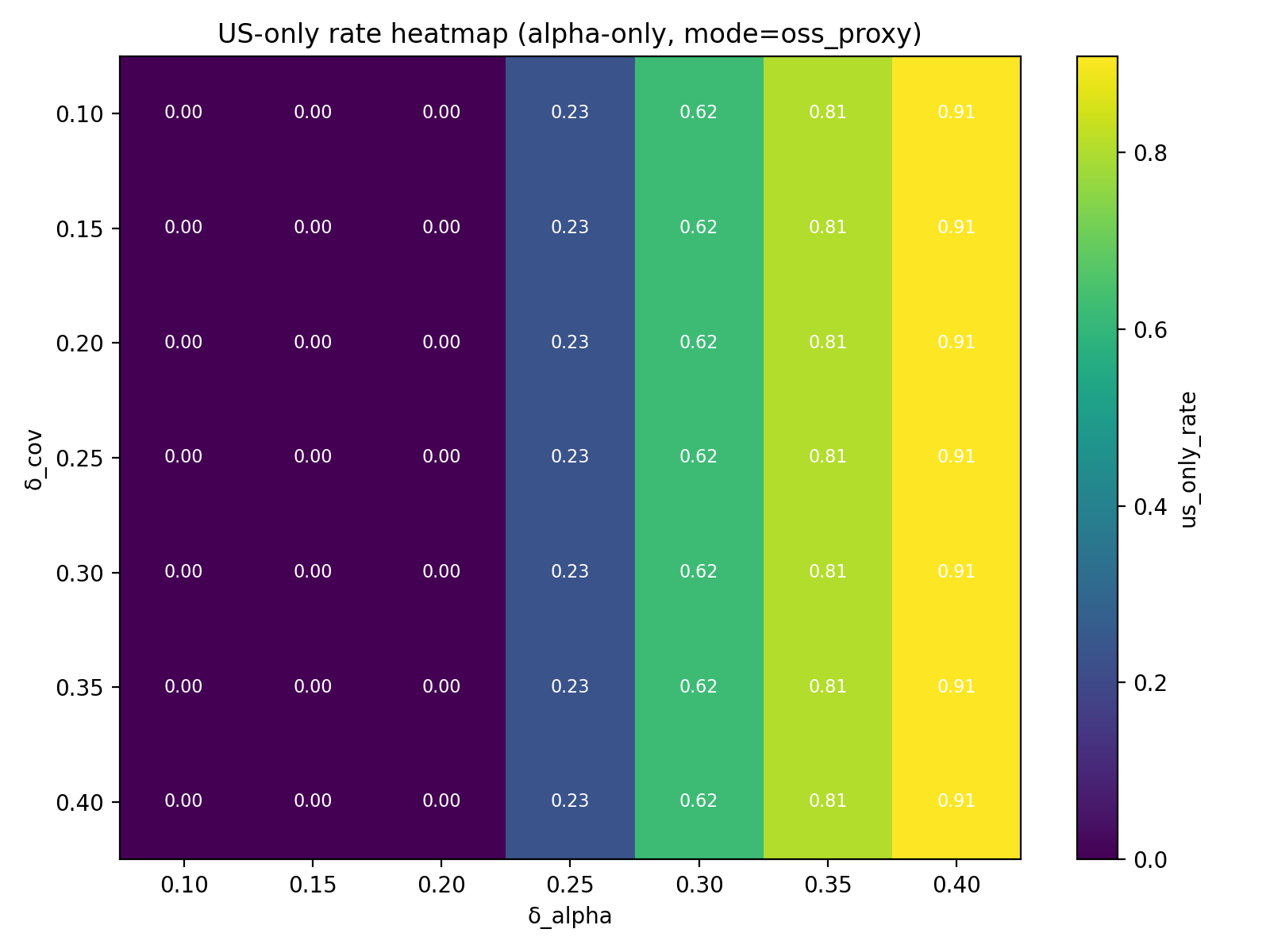}
\includegraphics[width=0.32\linewidth]{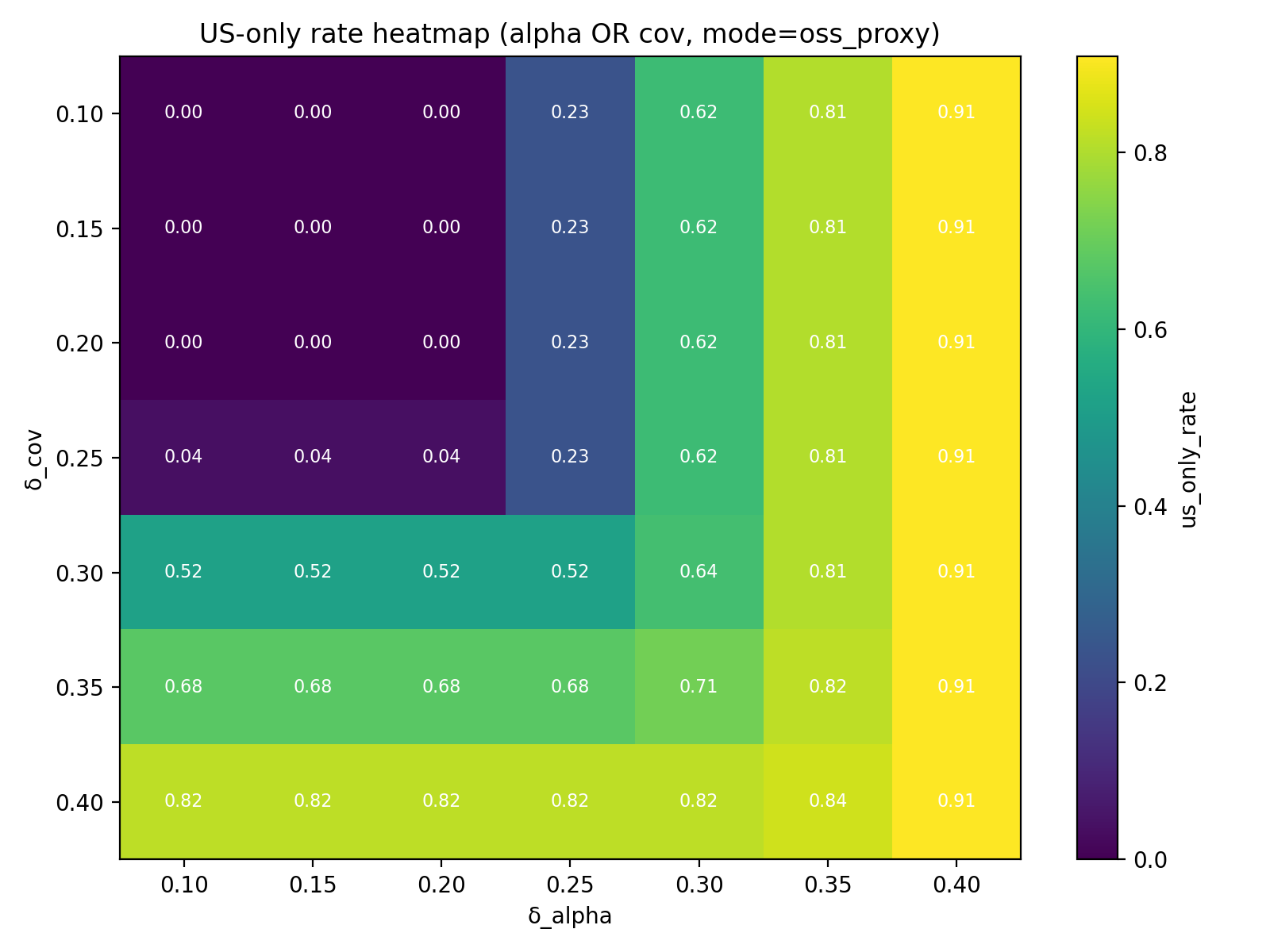}
\includegraphics[width=0.32\linewidth]{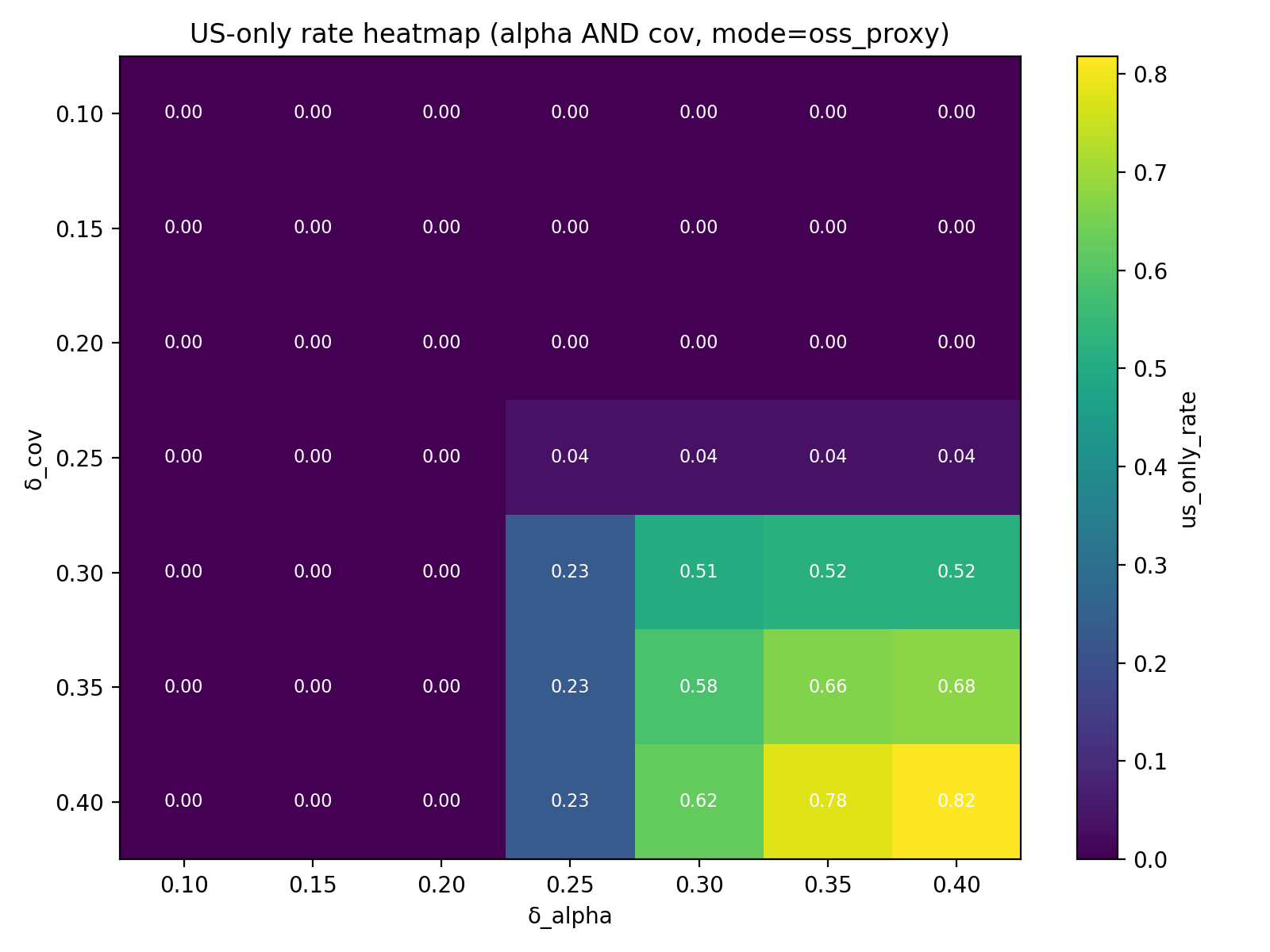}
\caption{US-only rate across miscoverage levels $(\delta_\alpha,\delta_{\mathrm{cov}})$ (target coverages $1-\delta$) for the three policy families. Smaller $\delta$ is more conservative.}
\label{fig:usonly-heatmaps}
\end{figure*}

\begin{figure*}[t]
\centering
\includegraphics[width=0.32\linewidth]{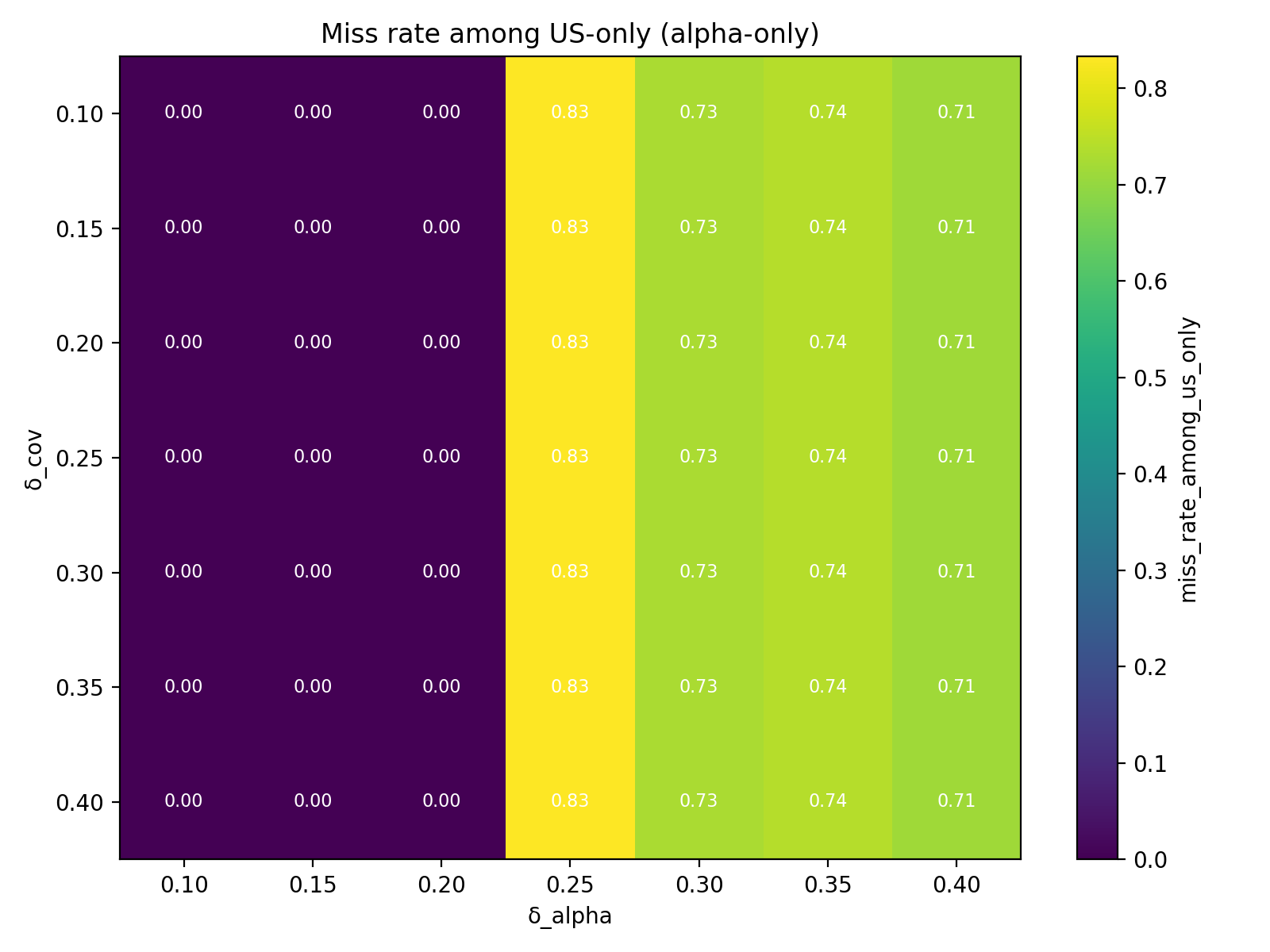}
\includegraphics[width=0.32\linewidth]{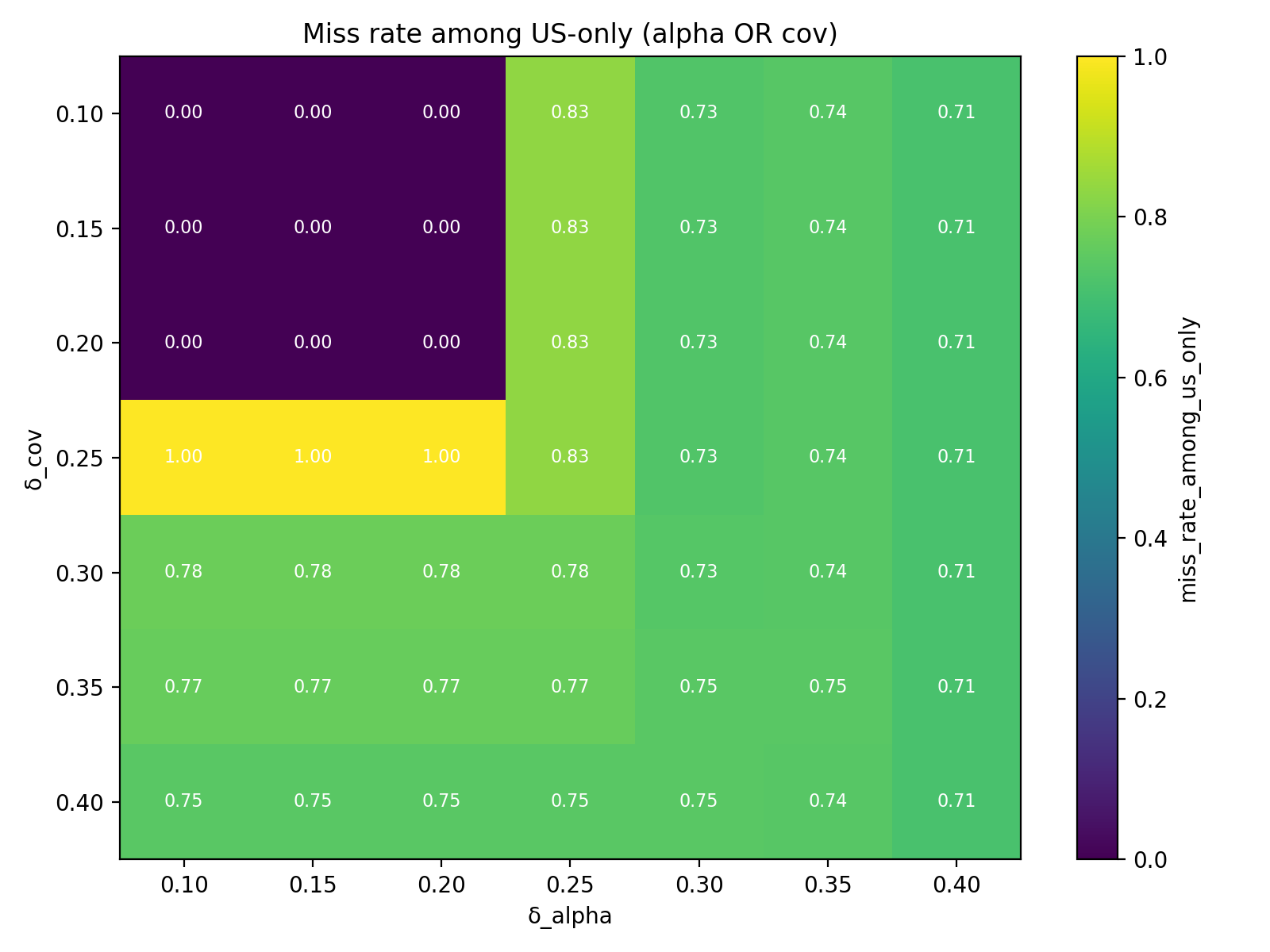}
\includegraphics[width=0.32\linewidth]{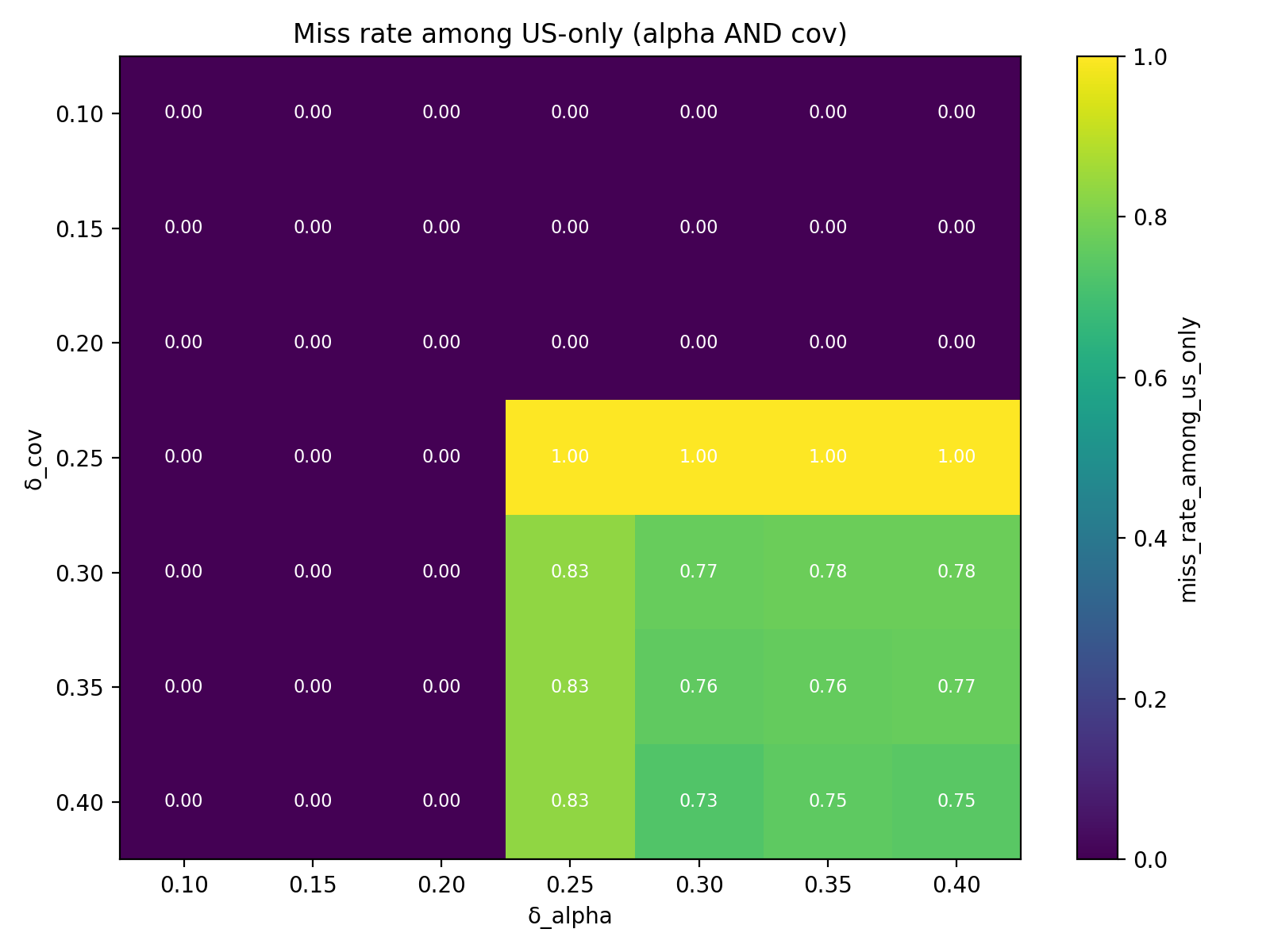}
\caption{Miss rate among US-only decisions across the same miscoverage grid $(\delta_\alpha,\delta_{\mathrm{cov}})$.}
\label{fig:miss-heatmaps}
\end{figure*}

\subsection{Calibration: affine bias correction and one-sided conformal bands}
For selective imaging we require calibrated lower bounds on US measurements with finite-sample (marginal) coverage under exchangeability. These bounds are then used inside the US-only decision rules. We perform two steps on the US calibration set $\mathcal{C}=\{(x_i,y_i)\}$ for each target $t\in\{\alpha,\mathrm{cov}\}$.

\paragraph{(i) Affine bias correction.}
Let $\hat{y}^t_i$ be the head prediction on $x_i$. We fit a robust affine correction
\[
\begin{aligned}
\tilde{y}^t_i \;=\; a_t\,\hat{y}^t_i \;+\; b_t\\
\text{by}\quad
(a_t,b_t) \;&\in\; \arg\min_{a,b}\ \sum_{(x_i,y_i)\in\mathcal{C}} \big|a\,\hat{y}^t_i + b - y^t_i\big|.
\end{aligned}
\]
This reduces small systematic bias without re-training the head.

\paragraph{(ii) One-sided residual quantiles.}
Define residuals $r^t_i = y^t_i - \tilde{y}^t_i$. For a desired miscoverage level $\delta_t\in(0,1)$ we compute a one-sided conformal radius
\[
q^+_t(\delta_t)\;=\;\mathrm{Quantile}_{\,\lceil (|\mathcal{C}|+1)(1-\delta_t)\rceil/|\mathcal{C}|}\Big(\,-r^t_i\ \text{over }(x_i,y_i)\in\mathcal{C}\Big),
\]
so that, under exchangeability, with probability $\ge 1-\delta_t$ a fresh sample will satisfy
$y^t \ge \tilde{y}^t(x) - q^+_t(\delta_t)$.
We then define the calibrated lower bound
\[
\mathrm{LB}_t(x;\delta_t) \;=\; \tilde{y}^t(x) \;-\; q^+_t(\delta_t).
\]
We sweep $\delta_\alpha$ and $\delta_{\mathrm{cov}}$ on a discrete grid (e.g., $0.10{:}0.40$) to produce coverage--utilization curves; smaller $\delta$ corresponds to higher target coverage ($1-\delta$) and more conservative bounds.

\subsection{Selective imaging rules (US-only vs.\ defer to XR)}
Let $T_\alpha$ and $T_{\mathrm{cov}}$ denote clinical normality thresholds (we use $T_\alpha{=}60^\circ$ and $T_{\mathrm{cov}}{=}50\%$). For an evaluation US image $x$, we declare “US-only” if the calibrated lower bounds exceed thresholds. We study three rule families:

\noindent\textbf{Alpha-only}
\begin{equation}
\label{eq:alpha-only}
d_{\alpha}(x) \;=\; \mathbb{I}\!\left[\mathrm{LB}_\alpha(x;\delta_\alpha)\ \ge\ T_\alpha\right].
\end{equation}

\noindent\textbf{Alpha OR Coverage}
\begin{multline}
\label{eq:alpha-or-cov}
d_{\alpha\vee\mathrm{cov}}(x) \;=\; 
\mathbb{I}\!\left[\mathrm{LB}_\alpha(x;\delta_\alpha)\ \ge\ T_\alpha\ \ \lor\right.\\
\left.\mathrm{LB}_{\mathrm{cov}}(x;\delta_{\mathrm{cov}})\ \ge\ T_{\mathrm{cov}}\right].
\end{multline}

\noindent\textbf{Alpha AND Coverage}
\begin{multline}
\label{eq:alpha-and-cov}
d_{\alpha\wedge\mathrm{cov}}(x) \;=\;
\mathbb{I}\!\left[\mathrm{LB}_\alpha(x;\delta_\alpha)\ \ge\ T_\alpha\ \ \land\right.\\
\left.\mathrm{LB}_{\mathrm{cov}}(x;\delta_{\mathrm{cov}})\ \ge\ T_{\mathrm{cov}}\right].
\end{multline}

Here $d(\cdot)\in\{0,1\}$ indicates “US-only” (1) vs.\ “defer to XR” (0). Sweeping $(\delta_\alpha,\delta_{\mathrm{cov}})$ yields a grid of policies.

\paragraph{Ossification-proxy variant.}
As a pragmatic sensitivity analysis, we also evaluate a variant that conditions thresholds on an ossific-nucleus flag $o\in\{0,1\}$ (from the US point annotation), e.g.,
\[
T_\alpha(o) = \begin{cases}
60^\circ,& o=0\\
60^\circ,& o=1
\end{cases}
T_{\mathrm{cov}}(o) = \begin{cases}
50\%,& o=0\\
50\%,& o=1
\end{cases}
\]
(kept equal in this study, but the machinery supports $o$-specific thresholds). This lets future work explore age/ossification-aware tuning without altering the conformal recipe.

\subsection{Pair construction and XR ground-truth events}
To align US decisions with XR outcomes, we form strict pairs on the evaluation split using $(\text{subject},\text{date},\text{side})$. For each pair $j$, define a binary XR abnormality indicator $z_j\in\{0,1\}$ that aggregates available radiographic ground truth (e.g., thresholding AI or CE when present, or using IHDI grades). We report all computations both per-rule and per-$(\delta_\alpha,\delta_{\mathrm{cov}})$ on the subset with at least one XR label.

\subsection{Empirical coverage and safety diagnostics}
On the evaluation US set $\mathcal{E}$ we compute empirical one-sided coverage for each target,
\[
\widehat{\mathrm{cvg}}_t
= \frac{1}{|\mathcal{E}|}\sum_{(x,y)\in\mathcal{E}} \mathbb{I}\!\left[y^t \ge \mathrm{LB}_t(x;\delta_t)\right],
\quad t\in\{\alpha,\mathrm{cov}\}.
\]
We also report miss rate among US-only decisions, $\mathrm{MR} = \frac{\sum_j d_j\,z_j}{\sum_j d_j}$, and the US-only rate, $\mathrm{UOR}=\frac{1}{N}\sum_j d_j$, where $N$ is the number of strict pairs.

\subsection{Decision-curve analysis (utility over cost-penalty grids)}
To summarize policy desirability across clinical preferences, we define a simple per-pair utility with radiation cost $\lambda\ge 0$ and miss penalty $\mu\ge 0$:
\[
u_j(d_j;\lambda,\mu,z_j) \;=\; -\,\lambda\,(1-d_j)\ -\ \mu\,z_j\,d_j.
\]
Acquiring XR ($d_j{=}0$) incurs cost $\lambda$; skipping XR ($d_j{=}1$) risks a penalty $\mu$ if the case is XR-abnormal ($z_j{=}1$). The average utility for a policy family $\Pi$ over $(\delta_\alpha,\delta_{\mathrm{cov}})$ and strict pairs $\{1,\dots,N\}$ is
\[
U_\Pi(\lambda,\mu) \;=\; \max_{(\delta_\alpha,\delta_{\mathrm{cov}})\in\mathcal{G}} \frac{1}{N}\sum_{j=1}^N u_j\!\big(d_j(\delta_\alpha,\delta_{\mathrm{cov}});\lambda,\mu,z_j\big),
\]
where $\mathcal{G}$ is the swept grid. We plot $U_\Pi(\lambda,\mu)$ versus $\lambda$ for fixed $\mu$ and compare against “acquire-all” and “acquire-none” baselines; this is analogous to net-benefit curves and exposes the radiation-safety trade-off transparently.

\subsection{Implementation details}
All images were resized to $512{\times}512$ with aspect-preserving padded crops; grayscale channels were replicated to three to match the ResNet stem. We trained SimSiam for 10 epochs per modality. For supervised heads, we used MAE for angle/coverage targets and cross-entropy for IHDI when labels were present; heads were trained on \emph{post-train} only using a fixed schedule and evaluated on \emph{evaluation}. The \emph{calibration} split is reserved for affine bias correction and conformal calibration. Encoders remained frozen throughout, aligning with the limited-label regime. Conformal calibration swept miscoverage levels $\delta_\alpha,\delta_{\mathrm{cov}} \in \{0.10,0.15,0.20,0.25,0.30,0.35,0.40\}$ (target coverages $1-\delta_\alpha$ and $1-\delta_{\mathrm{cov}}$). We report example radii for $\delta_\alpha=\delta_{\mathrm{cov}}=0.10$ in Results. Pair construction used strict (subject, date, side) matching; all reported pairwise metrics are computed on the subset with XR ground truth present.

\section{Results}

\begin{figure}[t]
\centering
\includegraphics[width=\linewidth]{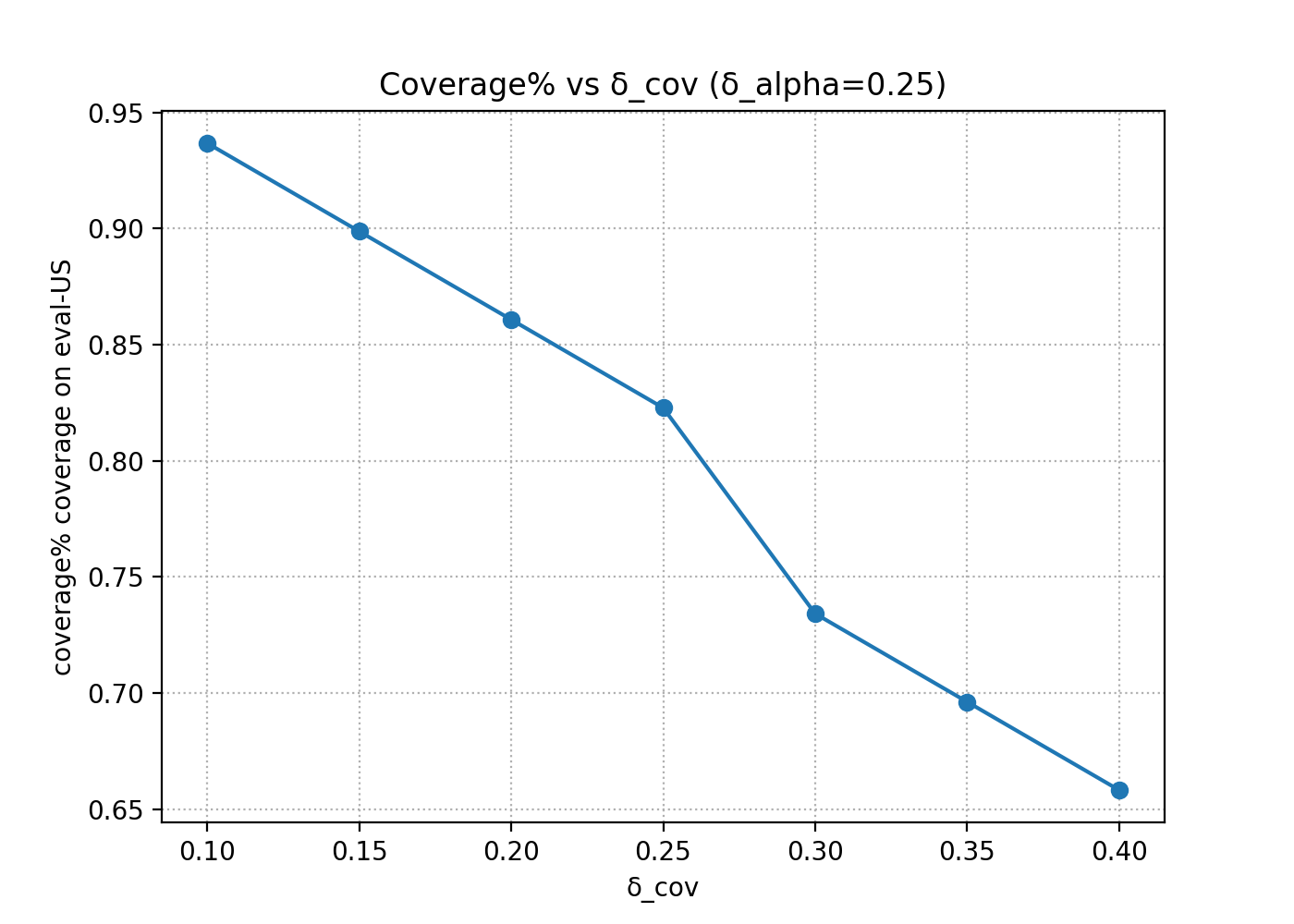}
\caption{Empirical one-sided coverage for femoral head coverage versus miscoverage $\delta_{\mathrm{cov}}$ (target coverage $1-\delta_{\mathrm{cov}}$).}
\label{fig:cov-lines}
\end{figure}

\begin{figure}[t]
\includegraphics[width=\linewidth]{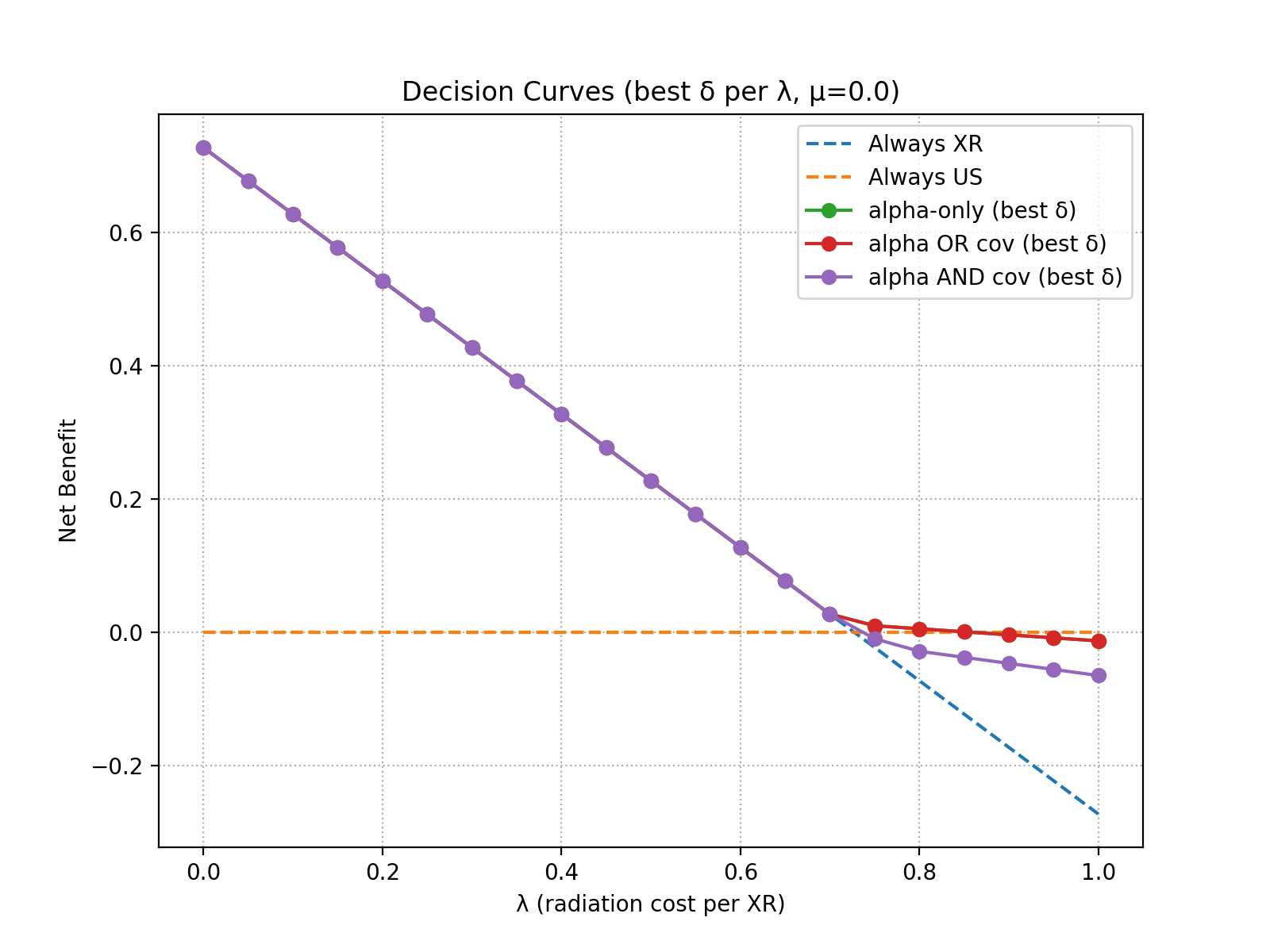}
\caption{Decision-curve envelopes over radiation cost $\lambda$ for miss penalty $\mu{=}0.0$. When $\mu$ is low, ultrasound-only becomes higher utility when radiation cost $\lambda$ is high. On the strict paired evaluation set, the OR policy (red) has higher utility than the AND policy (purple) at high radiation cost levels.}
\label{fig:dc}
\end{figure}

\begin{table}[t]
\centering
\caption{Policy snapshots (strict eval pairs, $N{=}77$).}
\begin{tabular}{lccc}
\hline
Rule & miscoverage $\delta_\alpha$/$\delta_{\mathrm{cov}}$ & US-only & XR use  \\
\hline
AND & 0.10 / 0.10 & 0.00 & 1.00 \\
AND & 0.20 / 0.20 & 0.00 & 1.00 \\
OR  & 0.35 / 0.35 & 0.43 & 0.57  \\
OR  & 0.40 / 0.40 & 0.55 & 0.45  \\
\hline
\end{tabular}
\label{tab:policy-snapshot}
\end{table}

\subsection{Probe Accuracy on Held-Out Evaluation}
Frozen encoders with small measurement heads achieved low double-digit mean absolute error (MAE) on ultrasound and single-digit MAE on radiographs. On the strict evaluation split, ultrasound probes yielded $\alpha$ MAE $= 9.69^\circ$, $\beta$ MAE $= 11.25^\circ$, and head-coverage MAE $= 13.97$ percentage points. Radiographic probes achieved acetabular index (AI) MAE $= 7.60^\circ$ and center-edge (CE) MAE $= 8.93^\circ$. These accuracies are competitive with unimodal reports that use larger bespoke networks, and they are achieved with frozen backbones and limited labels.

\subsection{Conformal Calibration on Ultrasound}
On the calibration split (7 subjects; 26 ultrasound images), we fitted per-target affine bias corrections then computed one-sided residual quantiles at miscoverage $\delta_\alpha=\delta_{\mathrm{cov}}=0.10$ (target coverage $0.90$). For $\alpha$ we obtained $q_\alpha^{+}{=}10.75^\circ$; for coverage, $q_{\mathrm{cov}}^{+}{=}28.74$ percentage points. Calibration MAEs were $\approx 6.4^\circ$ ($\alpha$) and $\approx 14.9$ percentage points (coverage). These radii define conservative lower bounds used by the policies in the next sections.

\subsection{Selective Imaging Policies: Coverage and Throughput}
We evaluated three ultrasound-first deferral rules on the strict paired set ($N{=}77$ hips with any XR ground truth), sweeping conformal miscoverage levels  $\delta_\alpha,\delta_{\mathrm{cov}}\in\{0.10{:}0.40\}$ (target coverage $1-\delta$):

\noindent\textbf{Alpha-only} requests an X-ray unless the calibrated lower bound for $\alpha$ exceeds $60^\circ$.  
\textbf{Alpha OR coverage} requests an X-ray unless either the $\alpha$ lower bound exceeds $60^\circ$ or the coverage lower bound exceeds $50\%$.  
\textbf{Alpha AND coverage} requests an X-ray unless both lower bounds exceed their thresholds.

Heatmaps of US-only rate (Fig.~\ref{fig:usonly-heatmaps}) and miss rate among US-only decisions (Fig.~\ref{fig:miss-heatmaps}) expose the expected trade-offs. At conservative settings (e.g., $\delta_\alpha{=}\delta_{\mathrm{cov}}{=}0.10$), the AND rule keeps US-only rate near zero while reaching empirical coverage $\approx 0.90$ for $\alpha$ and $\approx 0.94$ for coverage on the evaluation US set. Permissive settings under the OR rule increase US-only throughput substantially (e.g., $\sim 0.43$ at $0.35/0.35$ and $\sim 0.55$ at $0.40/0.40$) with corresponding increases in miss-rate heatmap intensity.

\subsection{Coverage as a Function of Conservatism}
Figure~\ref{fig:cov-lines} shows empirical one-sided coverage for femoral head coverage as a function of $\delta_{\mathrm{cov}}$ (with $\delta_\alpha$ fixed to the grid median), confirming monotone behavior with respect to miscoverage: empirical coverage decreases as  $\delta_{\mathrm{cov}}$   increases (equivalently, decreasing  $\delta$ yields more conservative bounds and higher coverage). The same pattern holds for $\alpha$ (not shown to avoid redundancy).

\subsection{Decision-Curve Analysis}
We summarize the radiation-safety trade using decision-curve style utilities $U(\lambda,\mu)$ that penalize each radiograph with cost $\lambda$ and each missed XR abnormality among US-only cases with penalty $\mu$. Figure~\ref{fig:dc} shows decision-curve envelopes for miss penalty $\mu{=}0.0$: for each radiation cost $\lambda$, we select the best $(\delta_\alpha,\delta_{\mathrm{cov}})$ on our grid \emph{within each rule family}. When radiation is costly (higher $\lambda$), more permissive policies that skip XR more often (higher US-only) tend to yield higher utility; as $\lambda$ falls (XR becomes cheaper), more conservative operating points that defer to XR more frequently become preferable. (Larger $\mu$ values shift the optimum toward more conservative operating points; not shown here.)

\subsection{Policy Snapshots for Reviewer Orientation}
Table~\ref{tab:policy-snapshot} lists a few interpretable operating points. Under AND at $0.10/0.10$, US-only rate is zero, XR utilization is unity, and empirical coverage is high ($\sim 0.90$ for $\alpha$; $\sim 0.94$ for coverage). Under OR at $0.35/0.35$ and $0.40/0.40$, US-only throughput increases to $\sim 0.43$ and $\sim 0.55$ while XR utilization drops accordingly. These snapshots match the contour gradients in the heatmaps.

\section{Discussion}

\paragraph{From measurements to decisions.}
Our goal is not to replace radiographs with ultrasound, but to formalize when ultrasound is enough and when an X-ray has positive value of information. The policy operates on named measurements that clinicians already use in practice, then converts them into calibrated lower bounds relative to clinical thresholds. This keeps the model on familiar ground and makes every decision traceable to a small number of interpretable quantities.

\paragraph{Collaborative decision support, not automation.}
The framework is designed to collaborate with sonographers, radiologists, and pediatric orthopedists. Clinicians retain control of thresholds and operating points through two dials. First, the rule family encodes clinical preference about how to combine evidence from $\alpha$ and femoral head coverage. Second, the miscoverage levels $(\delta_\alpha,\delta_{\mathrm{cov}})$ translate appetite for risk into stricter or looser lower bounds: smaller $\delta$ targets higher coverage and yields more conservative bounds; larger $\delta$ is more permissive. Heatmaps in Figure~\ref{fig:usonly-heatmaps} and Figure~\ref{fig:miss-heatmaps} are meant to be used as interactive guides during service setup. Teams can choose conservative points that preserve coverage and defer frequently to XR, or permissive points that reduce XR utilization while accepting a controlled increase in miss risk among US-only cases.

\paragraph{Radiation minimization as an explicit objective.}
Decision curves in Figure~\ref{fig:dc} make the radiation trade-off explicit. The utility weights $(\lambda,\mu)$ are not tuned by the model. They are handles for local policy. Sites that place a high cost on radiation (higher $\lambda$) will favor more permissive regions where US-only rates are higher; sites that place a lower cost on radiation (lower $\lambda$) can choose more conservative operating points with higher XR utilization. Sites that can tolerate a small increase in miss risk will move along the frontier to OR-rule settings that reduce X-ray utilization. This separates modeling from policy and puts radiation stewardship in the hands of the clinical team.

\paragraph{Safety properties and fail-safe behavior.}
The policy fails safe in two ways. First, one-sided conformal bounds give finite-sample (marginal) coverage control for the one-sided lower bounds. Empirical coverage trends in Figure~\ref{fig:cov-lines} reflect the intended monotonicity as miscoverage level increases. Second, cases near the thresholds route to XR. The pairwise analysis shows that many borderline hips sit within a few degrees of the $60^\circ$ alpha cutoff. Rather than stretching ultrasound to decide these, the system seeks a radiograph. This mirrors how experienced clinicians act under uncertainty.

\paragraph{Interpretability and auditability by design.}
The pipeline keeps the entire reasoning chain visible: line and point annotations, derived measurements, calibrated lower bounds, rule evaluation, and policy outcome. There are no hidden logits or opaque multi-class outputs gating safety-critical actions. This supports prospective quality assurance and after-action review. If a site wants to adjust a threshold, reweight the utility, or adopt ossification-aware criteria, the change reads like a guideline, not a re-training recipe.

\paragraph{How clinicians can use the figures.}
The three figure families provide complementary views for deployment. Heatmaps of US-only rate map operational throughput across $(\delta_\alpha,\delta_{\mathrm{cov}})$. Miss-rate heatmaps bound downside risk on the subset of US-only decisions. Decision curves summarize net benefit across radiation costs and highlight policy dominance regions. Together they provide a conservative floor and a permissive ceiling, giving a spectrum of defensible operating points that align with local practice and regulatory constraints.

\paragraph{Strengths relative to prior art.}
Most prior work treats ultrasound and radiography in isolation, often with black-box classifiers. Our approach uses measurement-faithful heads on both modalities, aligns decisions with long-standing clinical thresholds, and introduces coverage-controlled selective prediction to turn a unimodal estimate into a cross-modal action. This retains clinical vocabulary, supports mixed cohorts with variable ossification, and exposes an explicit safety knob.

\paragraph{Generalization and subgroup reliability.}
Real deployments will face device heterogeneity, operator variability, and demographic shifts. The conformal layer can be recalibrated per site, per device family, or per age band. Subgroup coverage audits can detect drift and trigger automatic tightening (decreasing) of the miscoverage levels $\delta$. Because the policy is measurement-based, these recalibrations are lightweight and do not require end-to-end re-training.

\paragraph{Clinical workflow integration.}
A practical path is ultrasound-first triage at the point of care. The model computes $\hat{\alpha}$ and $\widehat{\mathrm{cov}}$, applies bias correction, and renders calibrated lower bounds with a simple traffic-light presentation: green for US-only above threshold, gray for defer to XR, and an explanation panel showing the margin to threshold and the chosen $(\delta_\alpha,\delta_{\mathrm{cov}})$. Radiographs, when acquired, feed back into the registry and enlarge future calibration pools. This supports continuous learning while keeping the clinician in the loop.

\paragraph{Future work.}
Three extensions are natural. First, add modality-specific quality gates and age or ossification-aware thresholds so the rule respects validity domains automatically. Second, explore US$\rightarrow$XR surrogates that predict radiographic AI or IHDI risk from ultrasound measurements, then fold those calibrated risks into the decision curves. Third, run a prospective study that measures radiation saved per 100 infants, change in return-visit rates, and agreement with multidisciplinary adjudication. Each of these fits within the current policy scaffold without sacrificing interpretability.

\paragraph{Conclusion.}
This study shows that a small set of measurement-faithful probes, a calibration step that controls coverage, and a transparent policy grid are sufficient to build a collaborative assistant for selective imaging in DDH. The assistant is tunable to local radiation preferences, deferential near uncertainty, and legible to the clinicians who must own the decision. The figures and tables are intended as handrails for clinical setup and as a reproducible template for future multi-site validation.

% Uncomment the following to link to your code, datasets, an extended version or similar.
% You must keep this block between (not within) the abstract and the main body of the paper.
% \begin{links}
%     \link{Code}{https://aaai.org/example/code}
%     \link{Datasets}{https://aaai.org/example/datasets}
%     \link{Extended version}{https://aaai.org/example/extended-version}
% \end{links}

\bibliography{aaai2026}

% Check whether the conference requires a reproducibility checklist to be included in the paper.
% If so, you can uncomment the following line and ajust the path to include it.
% \input{../../ReproducibilityChecklist/LaTeX/ReproducibilityChecklist.tex}

\end{document}